\documentclass{article}
\usepackage{spconf,amsmath,graphicx}

\usepackage[utf8]{inputenc} 
\usepackage[T1]{fontenc}    
\usepackage{hyperref}       
\usepackage{url}            
\usepackage{booktabs}       
\usepackage{amsfonts}       
\usepackage{nicefrac}       
\usepackage{microtype}      
\usepackage{amsmath, amssymb, bbm, mathtools}
\usepackage{algorithm}
\usepackage[noend]{algpseudocode}
\usepackage{subcaption}
\usepackage{graphicx}
\usepackage{color}
\usepackage{cite}

\usepackage[nomargin,inline,draft]{fixme}
\fxsetup{theme=color,mode=multiuser}
\FXRegisterAuthor{hpm}{ahpm}{\color{green}hermina}
\FXRegisterAuthor{me}{ame}{\color{blue}mireille}
\FXRegisterAuthor{pf}{apf}{\color{red}pascal}
\FXRegisterAuthor{go}{ago}{\color{orange}Guille}
\FXRegisterAuthor{es}{aes}{\color{violet}Ersi}

\newcommand{\minimize}[2]{\ensuremath{\underset{\substack{{#1}}}{\mathrm{minimize}}\;\;#2 }}
\newcommand{\RR}{\mathbb{R}}

\newcommand{\corr}[1]{\textcolor{black}{#1}}

\makeatletter
\def\@name{\emph{Guillermo Ortiz-Jiménez}\qquad \emph{Mireille El Gheche}\qquad\emph{Effrosyni Simou}  \\ \emph{Hermina Petric Maretić}\qquad \emph{Pascal Frossard}\\}
\makeatother

\title{Forward-Backward Splitting for \corr{Optimal Transport based Problems}}
\address{École Polytechnique Fédérale de Lausanne (EPFL), Lausanne, Switzerland}
\begin{document}
%
\maketitle

\begin{abstract}
Optimal transport aims to estimate a transportation plan that minimizes a displacement cost. \corr{This is realized by optimizing the scalar product between the sought plan and the given cost, over the space of doubly stochastic matrices. When the entropy regularization is added to the problem, the transportation plan} can be efficiently computed with the Sinkhorn algorithm. Thanks to this breakthrough, optimal transport has been progressively extended to machine learning and statistical inference by introducing additional application-specific terms in the problem formulation. It is however challenging to design efficient optimization algorithms for optimal transport based extensions. To overcome this limitation, we devise a general forward-backward splitting algorithm based on Bregman distances for solving a wide range of optimization problems involving a differentiable function with Lipschitz-continuous gradient and a doubly stochastic constraint. We illustrate the efficiency of our approach in the context of continuous domain adaptation. Experiments show that the proposed method leads to a significant improvement in terms of speed and performance with respect to the state of the art \corr{for domain adaptation on a continually rotating distribution coming from the standard two moon dataset}.
\end{abstract}

\begin{keywords}
Forward-backward splitting, Bregman distance, optimal transport, continuous domain adaptation.
\end{keywords}

\section{Introduction}
\label{sec:intro}

Optimal transport (OT) is a fundamental notion in probability theory \cite{villani2008optimal} that defines a notion of distance between probability distributions by quantifying the optimal displacement of mass according to a so-called \textit{ground cost} associated to the geometry of a supporting space. The first formulation of optimal transport problem was introduced by Monge \cite{Monge_1781} in the 18th century, but it was later reformulated in a more tractable way by Kantorovich \cite{kantorovich1942transfer} in the early 1940s. However, the practical application of OT to solve computational tasks started only recently with the introduction of the entropic penalty to replace the nonnegative constraints in the transportation plan \cite{cuturi2013}. The resulting problem was significantly faster to solve than the unregularized counterpart using Sinkhorn algorithm \cite{sinkhorn_knopp_1967,Knight2008,cuturi2013}, \corr{which spread the use of OT for computing} distances between probability measures \cite{kolouri17,2016-peyre-icml,courty2016_gl,wang_deep_2018,MAL-073}.

Recent practice of OT has found application in multiple problems in signal processing and machine learning, such as color transfer \cite{rabin14}, image regestration\cite{Haker2004}, generative modelling\cite{arjovsky17a}, and domain adaptation\cite{courty2014_gl}. In this sense, most applications of OT work by extending the general optimal transport problem using a number of regularization terms \corr{and domain priors to structure the solution of the transport plan} \cite{Ferradans2014,Papadakis2014,courty2014_gl}. The resulting optimization problems can be solved using different algorithms. For example, if the loss function is differentiable, one can use the conditional gradient algorithm, which consists of linearizing the whole objective function \cite{Ber99,Jag13}. By linearizing only a part of the objective, the authors in \cite{rakoto2015} optimize a better approximation of the objective function, namely generalized conditional gradient splitting (CGS) algorithm. The latter however relies on line search to ensure the convergence, requiring an efficient solver of the partially linearized problem to be of strong interest. 

In this paper, we propose a new forward-backward splitting algorithm based on Bregman distances. The nice feature of the proposed method is that it works with a constant step-size, making it both efficient and easy to implement. \corr{To illustrate the flexibility of our approach, we develop an application of optimal transport to continuous domain adaptation}. We address the scenario in which the target domain is continually, albeit slowly, evolving, and in which, at different time frames, we are given a batch of test data to classify. This type of behavior can be seen in a variety of applications, such as traffic monitoring with gradually changing lightning and atmospheric conditions \cite{hoffman2014continuous}, spam emails evolving through time, or smooth regional variations of language across a country \cite{ruder2016towards}. Continuous domain adaptation has also found applications in healthcare, adapting the problem of X-ray segmentation to different domains \cite{venkataramani2018towards}. To the best of our knowledge, we are the first to tackle the problem of continuous domain adaptation using optimal transport. 

The remaining of the paper is organized as follows. Section \ref{sec:optimization} presents the general form of the optimization problem that we aim at solving and details the proposed forward-backward algorithm. Section \ref{sec:formulation} details the problem formulation of the the continuous domain adaptation. Section \ref{sec:experiments} provides results on a synthetic data for domain adaptation problem. Finally, Section \ref{sec:conclusions} concludes the paper.

\section{Optimization algorithm}
\label{sec:optimization}

\paragraph*{Problem setup.} Let $\mathcal{M}_{+}^{n}$ be the space of discrete probability measures of size $n$. Given two distributions $\mu^{{(1)}}\in \mathcal{M}_{+}^{n}$ and $\mu^{(2)}\in \mathcal{M}_{+}^{m}$, we denote by $C(\mu^{{(1)}}, \mu^{{(2)}}) \in \RR^{n\times m}$ to the transport cost between them, such that $[C]_{ij}$ measures the cost to transport one unit of mass from $[{\mu}^{(1)}]_i$ to $[\mu^{(2)}]_j$. We define the regularized optimal transport problem between these measures as 
\begin{align}\label{eq:prob_form}
\operatorname*{minimize}_{\gamma\in\mathbb{R}^{n\times m}} &\;\;\langle\gamma, C \rangle + \lambda H(\gamma)+J(\gamma)\\
\textrm{s.t.}&\quad \gamma \succcurlyeq 0,\;\; \gamma \mathbbm{1}^{m} =\mu^{(1)},\;\; \gamma^\top \mathbbm{1}^{n} = \mu^{(2)},\nonumber
\end{align}
where $H$ denotes the entropy operator\footnote{We define the entropy operator as
\begin{equation*}
H(\gamma) = \sum_{i,j} h(\gamma_{ij})
\quad\textrm{with }\\
h(\gamma_{ij}) = 
\begin{cases}
\gamma_{ij} \log\gamma_{ij} - \gamma_{ij} &\textrm{if $\gamma_{ij} > 0$}\\
0 &\textrm{if $\gamma_{ij} = 0$}\\
+\infty &\textrm{otherwise}.
\end{cases}
\end{equation*}}, $\lambda>0$, $\mathbbm{1}^{n} =[1\;\dots\;1] \in \mathbb{R}^{n}$, and $J\colon \mathbb{R}^{n\times m}\to\mathbb{R}$ is a differentiable function with $\beta$-Lipschitz continuous gradient that acts as regularizer for the transport plan. 


\paragraph*{Proposed algorithm} We propose to solve problem \eqref{eq:prob_form} via a forward-backward splitting algorithm based on Bregman distances \cite{van2017forward,combettes2019forward}. To this end, we remark that the Problem \eqref{eq:prob_form} can be generically formulated as
\begin{equation*}\label{eq:formulation}
\minimize{\gamma\in\mathcal{S}} \varphi(\gamma) + J(\gamma),
\end{equation*}
where $\varphi\colon \mathbb{R}^{n\times m}\to\mathbb{R}\cup\{+\infty\}$ is a lower semicontinuous convex function defined as
\begin{equation*}
	 \varphi(\gamma) =  \langle\gamma,C\rangle + \lambda H(\gamma),
	\end{equation*}
and $\mathcal{S} \subset \mathbb{R}^{n\times m}$ is a convex subset defined as
\begin{equation*}
	\mathcal{S} = \big\{ \gamma\in\mathbb{R}^{n\times m} \;|\; \gamma \succcurlyeq 0,\hspace{0.7em}\gamma \mathbbm{1}^{m} = \mu^{(1)},\; \gamma^\top \mathbbm{1}^{n} = \mu^{(2)}\big\}.
\end{equation*}
	%
	%
	
The above problem fits nicely into the forward-backward splitting framework of \cite{van2017forward,combettes2019forward}, which allows us to solve \eqref{eq:formulation} through the following iterative algorithm\footnote{$\iota_{\mathcal{S}}$ denotes the indicator function of $\mathcal{S}$, which is equal to $0$ for every $\gamma\in\mathcal{S}$, and $+\infty$ otherwise.}
\begin{equation}\label{eq:forward_backward}
	\gamma_{k+1} = \operatorname{prox}_{\alpha \varphi+\iota_{\mathcal{S}}}^{f}\big(\nabla f(\gamma_k) - \alpha \nabla J(\gamma_k)\big),
\end{equation}
where $\gamma_0 \in \mathbb{R}^{n\times m}$, $\alpha > 0$, and $f$ is a Legendre function. The key ingredient in the algorithm above is the $f$-proximity operator of $\varphi+\iota_{\mathcal{S}}$, which is defined as
\begin{equation*}
	\operatorname{prox}_{\alpha \varphi+\iota_{\mathcal{S}}}^{f}\big(\Sigma) = \operatorname*{argmin}_{\gamma \in \mathcal{S}}\; \alpha\varphi(\gamma) + f(\gamma) - \langle\gamma,\Sigma\rangle.
\end{equation*}
By setting $f=H$, the proximity operator boils down to an entropic optimal transport problem
\begin{equation*}
	\operatorname{prox}_{\alpha \varphi+\iota_{\mathcal{S}}}^{H}\big(\Sigma) = \operatorname*{argmin}_{\gamma \in \mathcal{S}}\; \langle\gamma,\alpha C - \Sigma\rangle + (1+\alpha\lambda) H(\gamma),
\end{equation*}
whose solution can be efficiently computed with the Sinkhorn algorithm \cite{cuturi2013}. According to the iterations in \eqref{eq:forward_backward}, replacing $\Sigma$ with $\nabla H(\gamma_k) - \alpha \nabla J(\gamma_k)$ leads to Algorithm \ref{alg:algo1}, which is guaranteed to converge to a solution to Problem \eqref{eq:formulation} by adequately setting the step-size $\alpha$, as discussed in \cite{van2017forward,combettes2019forward}.
	
\begin{algorithm}[t]
	\caption{Fast algorithm for the regularized optimal transport problem defined in  \eqref{eq:prob_form}.}
	\label{alg:algo1}
	\begin{algorithmic}[1]
		\Require{Function $J\colon \mathbb{R}^{n\times m}\to\mathbb{R}$ with $\beta$-Lipschitz continuous gradient}
		\Require{Cost $C\in\mathbb{R}^{n\times m}$, marginals $\mu^{(1)}\in\mathcal{M}_{+}^{n}$, and $\mu^{(2)} \in\mathcal{M}_{+}^{m}$} 
		\Require{Step-size $\alpha > 0$}
		\Require{Initialization $\gamma_0 \in]0,+\infty[^{n\times m}$}
		\For{$k=0,1,\dots\Big.$}
		\State $C_k = \alpha C + \alpha\nabla J(\gamma_k) - \log(\gamma_k)$
		\State $\gamma_{k+1} = \operatorname{sinkhorn}(C_k, 1+\alpha\lambda, \mu^{(1)}, \mu^{(2)})\Big.$
		\EndFor
		\State \textbf{return} $\gamma_{\infty}$ 
	\end{algorithmic}
\end{algorithm}

Note that Algorithm \ref{alg:algo1} is strikingly similar to the generalized gradient splitting algorithm (CGS) proposed in \cite{rakoto2015}. Indeed, they both consist of a sequential application of the Sinkhorn algorithm to an initial coupling, until it converges to a solution to the regularized problem. However, the CGS method performs a line search at each iteration to ensure the convergence, whereas Algorithm \ref{alg:algo1} simply works with a constant step-size, leading to an optimization method that is both \corr{more} efficient and much easier to implement \cite{van2017forward,combettes2019forward}.

\section{Continuous domain adaptation with optimal transport}
\label{sec:formulation} 

The vast majority of machine learning algorithms are designed and built around the assumption that the training and test samples are independent and identically distributed. Nevertheless, in some situations this is not the case, and in practice some distributional shift between the training and test distributions may cause a significant drop in the performance of the classifier. Domain adaptation algorithms \cite{wang_deep_2018} try to solve this mismatch, and propose ways to design classifiers that can handle differences in the test and trained distributions.

\begin{figure}
	\centering
	\includegraphics[width=\columnwidth]{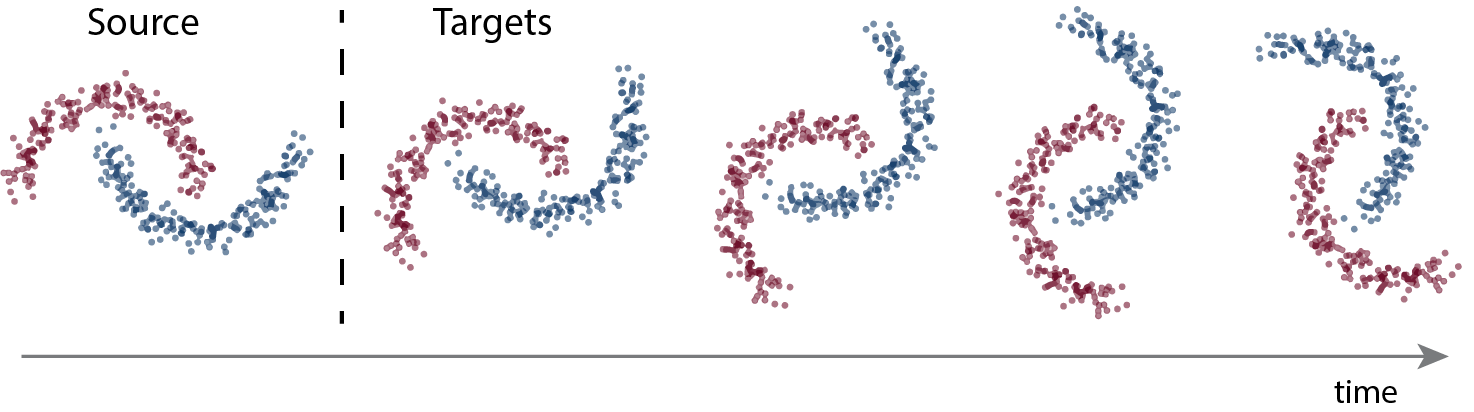}
	\caption{Example of a source domain and a sequence of slowly-varying target domains.}
	\label{fig:example}
\end{figure}

\corr{This happens when designing spam filters, for instance. Indeed, email features are in constant change, and a classifier trained to reject spam email in 2020 will not perform well in 2021. Similarly, a computer vision system deployed in an autonomous car would experience continuous changes in the atmospheric and lighting conditions that cause a continuous shift in the distribution of its inputs.} In this work, we propose to use a quadratic regularized optimal transport model \cite{Ferradans2014} to solve continuous domain adaptation problems \cite{hoffman2014continuous} for which the target domain is slowly evolving (cf. Figure \ref{fig:example}). To this end, we propose to extend the work in \cite{courty2016_gl}, and formulate a regularized optimal transport model that takes into account the transportation cost, the entropy of the probabilistic coupling, the labels of the source domain, and the similarity between successive target domains.
We denote the available discrete samples by $\{X^{(t)}\}_{t\in\mathbb{N}}$, where $X^{(0)}\in\RR^{n^{(0)}\times d}$ is the matrix of the source signal positions, and $X^{(t)}\in\RR^{n^{(t)}\times d}$ are the matrices of the moving target positions for each time $t\in\mathbb{N}$. In addition, we assume that the training samples $X^{(0)}$ are associated with a set of class labels $y^{(0)}\in \{1,\dots,L\}^{n^{(0)}}$, and the sequence of test samples are associated with unknown labels $\{y^{(t)}\}_{t>0}$.

In order to infer the unknown labels, we propose to estimate a sequence of signal mappings $T_t:\RR^{n^{(0)}\times d}\rightarrow\RR^{n^{(0)}\times d}$ that assign each source sample a position in the target domain. As is common in OT we define $T_t$ as the barycentryc mapping of the source signals $X^{(0)}$ to the target signal at time $t$
\begin{equation}
 	T_t(X^{(0)})=n^{(t)}\gamma^{(t)} \; X^{(t)}.
\end{equation} 
Under this parameterization, adaptation is performed by estimating the sequential transport plans $\{\gamma^{(t)}\}_{1\leq t \leq M}$ using a set of regularized optimal transport problems. In this setup, we will use $C^{(t,t+1)} \in \RR^{n^{(t)}\times n^{(t+1)}}$ to denote the transportation cost from source to target at time $t$, where each entry $[C^{(t,t+1)}]_{ij}$ contains the Euclidean distance between $[T_{t-1}(X^{(0)})]_{i,:}$ and $[X^{(t)}]_{j,:}$.

In short, to perform the continuous domain adaptation we do:
\begin{enumerate}
\item First, we compute the probabilistic coupling between the source distribution $\mu^{(0)}$ and the first target distribution $\mu^{(1)}$ as the solution to the entropic optimal transport
\begin{equation*}
\gamma^{(0)} = \operatorname*{argmin}_{\gamma\in\mathcal{S}_0} \langle \gamma,C^{(0,1)}\rangle + \lambda H(\gamma),
\end{equation*}
where, $\forall t \in\mathbb{N}$,
\begin{equation*}
    \mathcal{S}_t=\{ \gamma\in\mathbb{R}_+^{n\times m} |\gamma \mathbbm{1}^{n^{(t+1)}}  = \mu^{(t)},  \gamma^\top \mathbbm{1}^{n^{(t)}} = \mu^{(t+1)}\}. 
\end{equation*}

\item Then, for every $t\in\mathbb{N}\setminus{\{0\}}$, we compute the probabilistic coupling between the distribution $\mu^{(t)}$ and the subsequent distribution $\mu^{(t+1)}$ as follows
\begin{align}
\gamma^{(t)} &=  \operatorname*{argmin}_{\gamma\in\mathcal{S}_t}\; \langle \gamma,C^{(t,t+1)}\rangle + \lambda H(\gamma)\label{eq:cdot}\\
&\hspace{4.5em}+\eta_c R_{c}(\gamma)+\eta_t R_t(\gamma),\nonumber
\end{align}
where $\eta_c>0$, $\eta_t>0$, $R_c$ is a class-based regularizer, and $R_{t}$ is a time-based regularizer that promotes smoothness of the transport plan through time.

The class regularizer aims to convey label-based information that is grounded on the assumption that each target sample has to receive masses only from source samples that have the same label. In this work we follow the term proposed in \cite{courty2014_gl} and set
\begin{equation*}
R_c(\gamma)= \sum_{j}\sum_{\ell} \|\gamma(\mathcal{I}_\ell,j)\|_2.
\end{equation*}
Here above, $\mathcal{I}_\ell \subset \{1,\dots,n^{(t)}\}$ gathers the row indices of $\gamma\in \mathbb{R}^{n^{(t)}\times n^{(t+1)}}$ that belong to the same class $\ell\in\{1,\dots,L\}$. The mixed norm is used in order to model the ``group sparsity'', e.g., dependencies between the group of points that belong to the same class. 
 
The novelty of our work stems from the additional temporal regularization which is modeled via a smoothness penalization of the barycentric mapping based on 
\begin{equation*}
R_t(\gamma) = \|n^{(t)} \gamma \; X^{(t)} - n^{(t-1)}  \gamma^{(t-1)} X^{(t-1)}\|^2_F.
\end{equation*}

\item Finally, we train a classifier on the mapped source samples $n^{(t-1)}\gamma^{(t-1)} X^{(t-1)}$ and evaluate the accuracy on the new target datapoints $ X^{(t)}$.
	\end{enumerate}

\section{Experiments}
\label{sec:experiments}

\subsection{Speed of convergence of the optimization algorithm}
\begin{figure}[ht!]
	\centering
	\begin{subfigure}[b]{\columnwidth}
	\includegraphics[width=0.95\columnwidth]{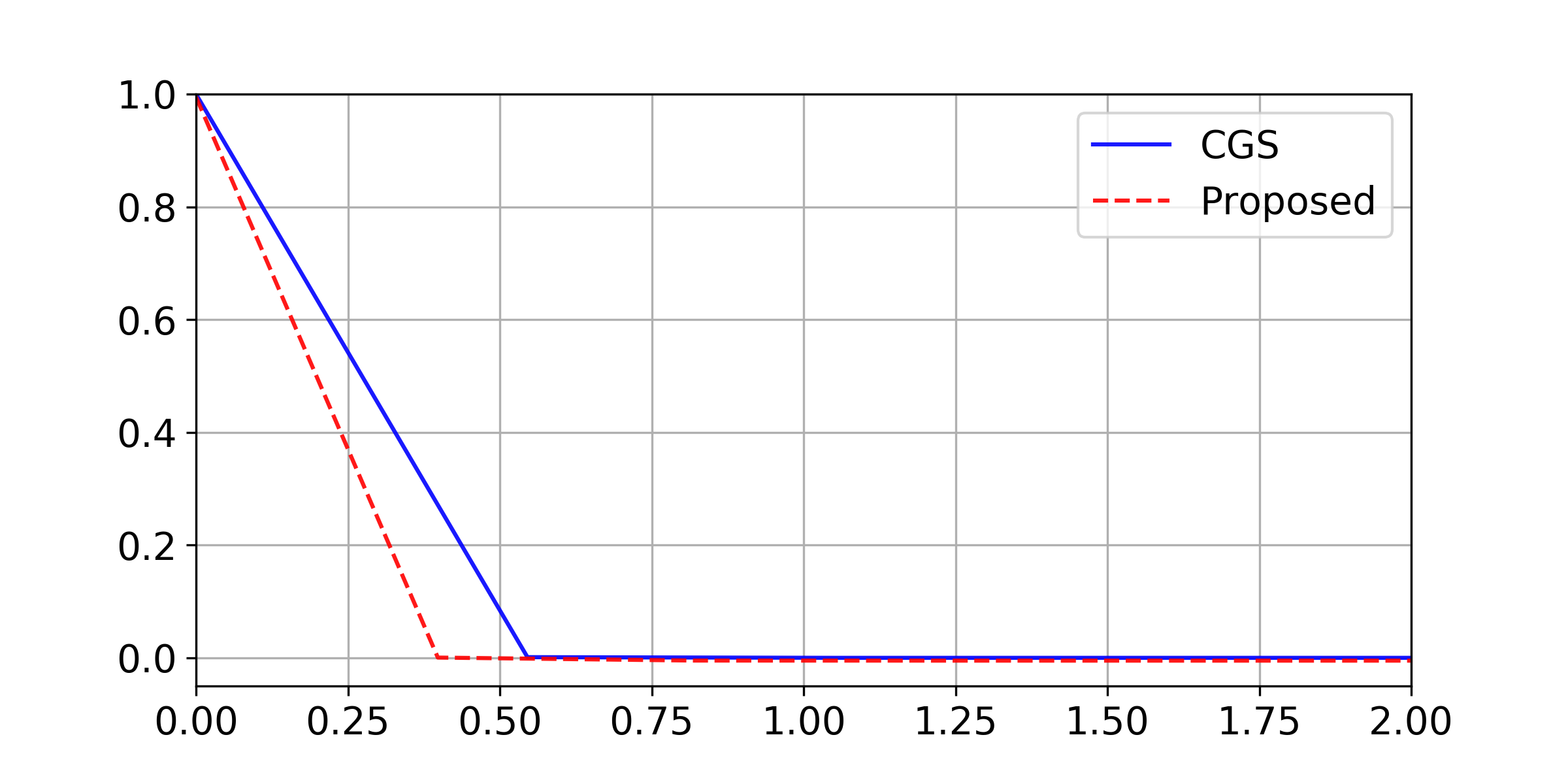} 
	\caption{$\lambda=0.5$, $\eta_c=0$, $\eta_t=50$.}
	\end{subfigure}
	\hfill 
	\begin{subfigure}[b]{\columnwidth}
	\includegraphics[width=0.95\columnwidth]{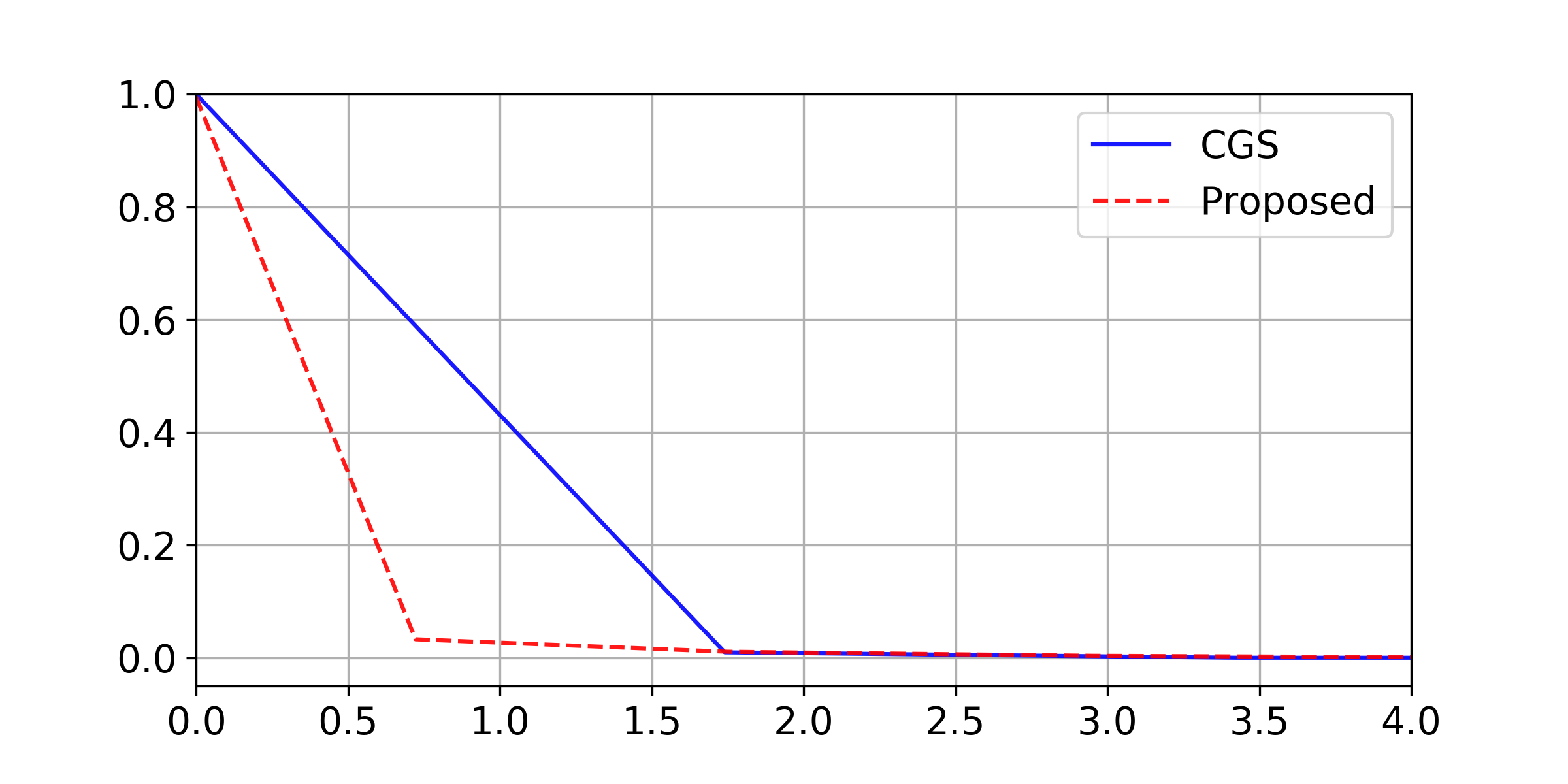}
	\caption{$\lambda=0.01$, $\eta
	_c=0$, $\eta_t=50$.}
	\end{subfigure}
	\caption{Objective value versus time. The step size is set to $\alpha=10$.}
	\label{fig:convergence}
	\vspace{-2em}
\end{figure}

\corr{We first study the performance of the proposed algorithm to solve the regularized adaptation step in \eqref{eq:cdot}. In particular, we compared the performance of our algorithm with CGS using two different sets of regularization parameters in a domain adaptation task with $5000$ samples in the source domain and $1000$ in the target domain taken from the standard two moon dataset.} For both Algorithm \ref{alg:algo1} and CGS algorithm \cite{rakoto2015}, Figure \ref{fig:convergence} reports the normalized cost evaluations versus the cumulative time per iteration. The curves show that the proposed approach converges faster than CGS, especially with a low entropic regularization (i.e., small $\lambda$). This is due to the fact that one iteration of Algorithm \ref{alg:algo1} is cheaper than one iteration of CGS algorithm, due to the line search performed by the latter to adjust the step size. 

\subsection{\corr{Continuous domain adaptation performance}}
To assess the effectiveness of the proposed time regularization, we compare the adaptation and tracking performance of several optimal transport strategies that use different combinations of regularizers to perform the domain adaptation. In our experiments we replicate the setup proposed in \cite{courty2016_gl} where they use the standard two entangled moon dataset as source. To create the sequence of targets we sample new data points from the source distribution and rotate them around the origin in batches with steps of $18$ degrees. In our simulations we use $500$ labeled data samples in the source domain, and $50$ samples in each target domain.  After each adaptation step, we train a new 1-Nearest Neighbor classifier on the mapped source samples and evaluate its accuracy on a $1000$ new datapoints for each target.

We compare three different methods for continuous domain adaptation that use optimal transport: The algorithm proposed in \cite{courty2016_gl}, where on top of the typical entropic regularization, they add a group lasso regularization term to the optimal transport problem to penalize transport mappings where samples from different classes in the source are coupled with the same samples in the target. Our proposed algorithm, in which we add the time regularization term introduced in Section~\ref{sec:formulation} to promote temporal smoothness. And finally, a combination of the two algorithms where we add both regularizers in the optimization. Besides, for each of the algorithms we run two sets of experiments: 
\begin{itemize}
    \item \textbf{Sequential cost (seq)}: We sequentially map the source samples $X^{(0)}$ to the targets at time $t\in\mathbb{N}$. We use the positions of the mapped samples $T_t(X^{(0)})$ and the positions of $X^{(t+1)}$ to compute the optimal transport cost $C^{(t,t+1)}\in \RR^{n^{(t)}\times n^{(t+1)}}$. 
    
    \item \textbf{Static cost}: We fix the source samples to $X^{(0)}$ and directly match them to the target samples $X^{(t)}$ at time $t\in\mathbb{N}$. In this case, the transport cost $C^{(0,t)}\in \RR^{n^{(0)}\times n^{(t)}}$.
\end{itemize}
\corr{In each set of experiments, we compare three different settings: Using only time-based regularization, only class based-regularization and a combination of the two. The hyperparameters in each setting are optimized using grid search on a validation test different than the one used for testing.}

\begin{figure}[ht!]
	\centering
	\includegraphics[width=0.9\columnwidth]{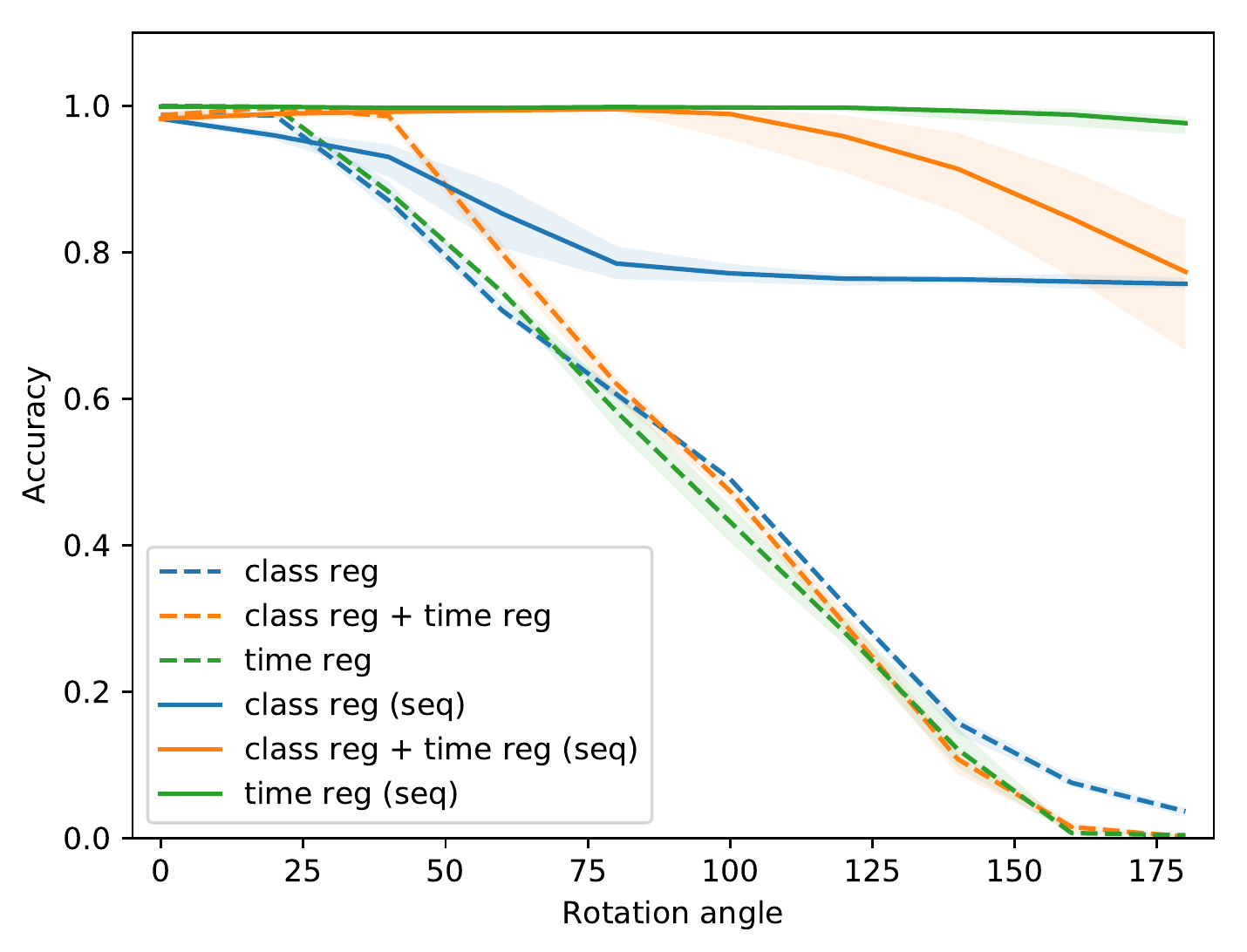}
	\caption{Performance comparison of different continuous domain adaptation strategies with optimal transport. Plot shows average and minimum and maximum values over 10 runs using the best regularization parameters for each of the methods (tuned using grid search on different samples).}
	\label{fig:moons}
	\vspace{-1.5em}
\end{figure}

Figure~\ref{fig:moons} shows the performance of the different methods. Clearly, the use of a sequential adaptation strategy, instead of a static one, allows for better tracking and adaptation. Furthermore, we can see that using the previously proposed group lasso regularization on the source labels\cite{courty2014_gl,courty2016_gl} is not enough to guarantee a continuous adaptation. On the contrary, the time regularizer ensures temporal consistency along the sequence of adaptations and preserves the accuracy of the classification method on all the targets.
\section{Conclusions}
\label{sec:conclusions}

\corr{We have presented an efficient algorithm to solve regularized optimal transport problems based on forward-backward splitting and Bregman divergences. In the most general form, this algorithm can be used to minimize any differentiable function with Lipschitz-continuous gradients and doubly stochastic constraints. This algorithm is more efficient and easier to implement than the competing CGS. Additionally, we have introduced a new optimal transport framework for continuous domain adaptation on slowly varying domains. Our solution is based on the introduction of a temporal regularization term in the optimal transport problem that promotes smoothness along the trajectory of the mapped source samples.} Finally, we have tested our framework on a synthetic example and showed its superior performance over the state-of-the-art algorithms. In future work, we plan to extend our temporal regularization term to different metrics like the Wasserstein distance, and to work on accelerated versions of our optimization algorithm.

\bibliographystyle{IEEEbib}
\bibliography{strings,refs}

\end{document}